\pdfoutput=1

\documentclass[11pt]{article}

\usepackage[]{acl}

\usepackage{times}
\usepackage{latexsym}

\usepackage[T1]{fontenc}

\usepackage[utf8]{inputenc}

\usepackage{microtype}

\usepackage{graphicx}
\usepackage{booktabs}       
\usepackage{amsfonts}       
\usepackage{nicefrac}
\usepackage{latexsym}
\usepackage{multirow}
\usepackage{amsmath}
\usepackage{arydshln}
\usepackage{subfigure}
\usepackage{colortbl}
\usepackage{hyperref}

\title{ \textsc{CoLo}: A Contrastive Learning based Re-ranking Framework for One-Stage Summarization}

\definecolor{mygray}{gray}{0.9}
\def\best{\cellcolor[gray]{0.9}}

\author{Chenxin An$^{1}$, Ming Zhong$^2$, Zhiyong Wu$^3$  ,Qin Zhu$^1$, Xuanjing Huang$^1$, Xipeng Qiu$^{1}$\thanks{\ \  Corresponding author.}  \\
  $^1$School of Computer Science, Fudan University \\
  $^2$ University of Illinois at Urbana-Champaign \\
  $^3$ Shanghai AI Lab \\
 \texttt{\{cxan20, qzhu18, xjhuang, xpqiu\}@fudan.edu.cn }\\
 \texttt{ mingz5@illinois.edu, wuzhiyong@pjlab.org.cn}
  }

\begin{document}
\maketitle
\begin{abstract}
Traditional training paradigms for extractive and abstractive summarization systems always only use token-level or sentence-level training objectives.
However, the output summary is always evaluated from summary-level which leads to the inconsistency in training and evaluation.
In this paper, we propose a \textbf{Co}ntrastive \textbf{L}earning based re-ranking framework  for \textbf{o}ne-stage  summarization called \textsc{CoLo}. By modeling a contrastive objective, we show that the summarization model is able to directly generate summaries according to the summary-level score without additional modules and parameters.  Extensive experiments demonstrate that \textsc{CoLo} boosts the extractive and abstractive results of one-stage systems on CNN/DailyMail benchmark to 44.58 and 46.33 ROUGE-1 score while preserving the parameter efficiency and inference efficiency.
Compared with state-of-the-art multi-stage systems, we save more than 100 GPU training hours and obtaining \textbf{$3\times \sim 8\times$} speed-up ratio during inference while maintaining comparable results\footnote{\url{https://github.com/ChenxinAn-fdu/CoLo}}.

\end{abstract}

\section{Introduction}

In general, there are two main paradigms to do text summarization: \textit{abstractive}~\cite{rush2015neural,nallapati2016abstractive,gehrmann2018bottom} and \textit{extractive}~\cite{cheng2016neural,narayan2018ranking, zhong2019searching, zhong2022unsupervised} methods.

For extractive summarization, previous studies~\cite{nallapati2017summarunner, liu2019text} formulate it as a \textit{sentence-level} sequence labeling task.
However, there is an inherent gap between the \textit{sentence-level}  scoring and the \textit{summary-level} evaluation~\cite{DBLP:conf/acl/ZhongLCWQH20}.This means that some high-scoring sentences may share the same meaning, making them not a qualified summary when combined.
Similarly, the previous training paradigm for abstractive summarization models can be viewed as a \textit{token-level} scoring process upon the decoder of sequence-to-sequence model. There also exists the issue of exposure bias~\cite{bengio2015scheduled, paulus2017deep} in the teacher-forcing framework leading to the error accumulation during auto-regressive decoding. Therefore, previous frameworks for both extractive and abstractive methods did not perform \textit{summary-level} optimization. 

To tackle this problem, state-of-the-art summarization systems~\cite{DBLP:conf/acl/ZhongLCWQH20, liu2021simcls} are enhanced with an additional module (called re-ranker) and follow a two-stage paradigm. They first train a  summarizer to model the conditional distribution $p(Y|X)$ where $X$ is the document and $Y$ is the output summary. Then the re-ranker is trained to re-score candidates sampled from the pre-trained summarizer in the second stage. However, this paradigm trades efficiency for accuracy, the auxiliary re-ranking greatly harms the inference efficiency especially for the highly efficient extractive systems. Experimentally, the decoding speed of two-stage re-ranking models  is only \textasciitilde{{7.0}} samples/s while removing the re-ranker module will greatly boost the decoding speed to \textasciitilde{\textbf{42.0}} samples/s\footnote{We run these two models on the test set of CNN/DailyMail using single GeForce GTX TITAN XP GPU for 3 times and report the average speed.}. This makes two-stage summarization systems may be unacceptable in real-world scenarios that require timely feedback.

The limitations of the existing work motivate us to build a one-stage summarization system that can 1) replace previous naive sentence/token-level score with a summary-level score and 2) do not sacrifice the parameter and inference efficiency.
In this paper, we propose a  \textbf{Co}ntrastive \textbf{L}earning based re-ranking framework for \textbf{o}ne-stage summarization called \textsc{CoLo} for both extractive and abstractive approach. Contrastive learning has been explored in summarization~\cite{sun2021alleviating,an2021retrievalsum} and generation~\cite{lee2020contrastive, ancont}. \textsc{CoLo} uses a contrastive re-ranking training objective.

We first present a novel sampling method that can be equipped to any one-stage summarization systems so that it can re-score candidates without the second stage.

The existing two-stage models use \textbf{offline sampling} to preprocess samples for training of re-ranker where candidate samples are drawn from a fixed model distribution.

This is a huge obstacle to turning \textit{summarize-then-rerank} two-stage framework into an efficient end-to-end model.
To solve this issue, we propose an \textbf{online sampling} approach.
Concretely, instead of sampling from a fixed distribution, we draw positive and negative samples from a dynamic distribution of model outputs during training, which ultimately eliminates the requirement for additional modules in the overall framework.

We then introduce a summary-level optimization strategy in addition to the traditional sentence-level (for extractive systems) or token-level loss (for abstractive systems).
As a result, as a one-stage model, \textsc{CoLo} achieves comparable performance to two-stage systems, and greatly improves decoding speed to meet the needs of real-world applications.

We summarize our contributions as follows:
\begin{itemize}
    \item We are the first to propose a one-stage re-ranking framework \textsc{CoLo} for both extractive and abstractive summarization systems. 
    \item Results on the popular CNN/DailyMail benchmark show that both the extractive and abstractive versions of \textsc{Colo} outperform previous state-of-the-art one-stage systems by a large margin. Compared to the two-stage systems, \textsc{CoLo} achieves comparable performance without additional pre-trained model. More importantly, \textsc{Colo} do not sacrifice inference speed and thus can be more widely used in real-world scenarios.
\end{itemize}

\section{Background}
\subsection{Preliminary about Two-Stage Systems }
Two-stage paradigms~\cite{DBLP:conf/acl/ZhongLCWQH20,liu2021simcls} improve summarization quality by re-ranking and selecting a candidate from a given set of candidates. MatchSum~\cite{DBLP:conf/acl/ZhongLCWQH20} forms a contrastive learning based re-ranking framework where they first generate a set of candidates summaries by a extractive summarization model and then feed them to a re-ranker. The re-ranker is trained to optimize a summary-level score and it can evaluate the candidate summaries holistically.  SimCLS~\cite{liu2021simcls} is the abstractive version which replaces the extractive  summarizer in \citet{DBLP:conf/acl/ZhongLCWQH20} with a abstractive summarizer. 

The training objective for summarization models is to estimate a conditional probability distribution $p(Y|X)$, where $X$ is the document and $Y$ is the output summary.  Given a summarization model $\mathcal{M}$ that has already tuned under the conventional framework with loss function $\mathcal{L}_{sum}$ where $\mathcal{L}_{sum}$ could be binary cross entropy loss (BCELoss) or negative log likelihood loss (NLLLoss).  The two-stage systems should first use a sampling algorithm e.g. beam search to sample a candidate set $\mathcal{C} = \{C_1, C_2, \ldots, C_m\}$ of size $m$ from the fixed model distribution $C_i \sim p_{\mathcal{M}}(Y|X)$. Candidates in  $\mathcal{C}$ are sort by their ROUGE score in descending order.  Then the they further train a separate re-ranker,e.g., BERT , with a contrastive-style ranking loss $\mathcal{L}_{rank}$ to select the the best candidate from  $\mathcal{C}$ as the final output.   The ranking loss used in the best re-ranking system for summarization is the triplet margin loss~\cite{kingma2014adam}. For a candidate pair $(C_i, C_j)$ where $i<j$, if $C_i$ has higher ROUGE score and it will be treated as the positive sample:
\begin{equation}
    \mathcal{L}_{i,j} = max\{0, \text{cos}(\mathbf{z}_X, \mathbf{z}_{C_j})  -  \text{cos}(\mathbf{z}_X, \mathbf{z}_{C_i})  + \rho \},
\label{eq:loss_rank}
\end{equation}
where $\mathbf{z}_X, \, \mathbf{z}_{C_i},\, \mathbf{z}_{C_j}$ are  the vector feature representation of  $X, C_i, C_j$ output by  the re-ranker, and $\rho$ is the margin value.  The final ranking loss is obtained by summing up all pairs: $\mathcal{L}_{rank} = \sum_j\sum_{i<j} \mathcal{L}_{i,j}$. The ranking loss ensures that candidates with higher ROUGE score is closer to the document in the embedding space.

\subsection{A Comparison between Two-Stage Systems and \textsc{CoLo}}
Figure~\ref{fig:compare} illustrates the difference between the architecture of two-stage systems and \textsc{CoLo}. Although MatchSum and SimCLS significantly outperform all one-stage models, they mainly suffer from three drawbacks which strongly emphasize the necessity of designing an one-stage model: 

(1) Training/inference inefficiency. Building the training set of the re-ranker and the second training stage consumes large amounts of GPU and CPU time (see details in Section~\ref{sec:effi}).
Moreover, the need of re-feeding generation results to another module also requires unaffordable computational resources. 

(2) Coupling between the summarizer and re-ranker. Each improvement to one of these modules requires simultaneous updating or retraining of another module, which limits the use of such systems in the real world. For example, to try a larger candidate set or a different decoding method, we have to prepare the training set again for the second stage. In addition, how to tune the hyperparameters to be optimal in both modules at the same time is another tricky issue. Compared with two-stage systems, our one-stage system has a simple and clean implementation.

(3) Two-stage systems also face difficulties in long document summarization, because the input length of the re-ranker will drastically increase as the length of candidates increasing (see detailed analysis in Appendix~\ref{sec:long_doc}).
Correspondingly, \textsc{CoLo} is not easily affected by length variance.

\newcommand{\rulesep}{\unskip \vrule width 0.5pt}
\begin{figure*}[ht!]
  \centering

  \subfigure[Two-stage models: MatchSum and SimCLS]{
    \label{fig:rerank}
    \includegraphics[width=0.42\textwidth]{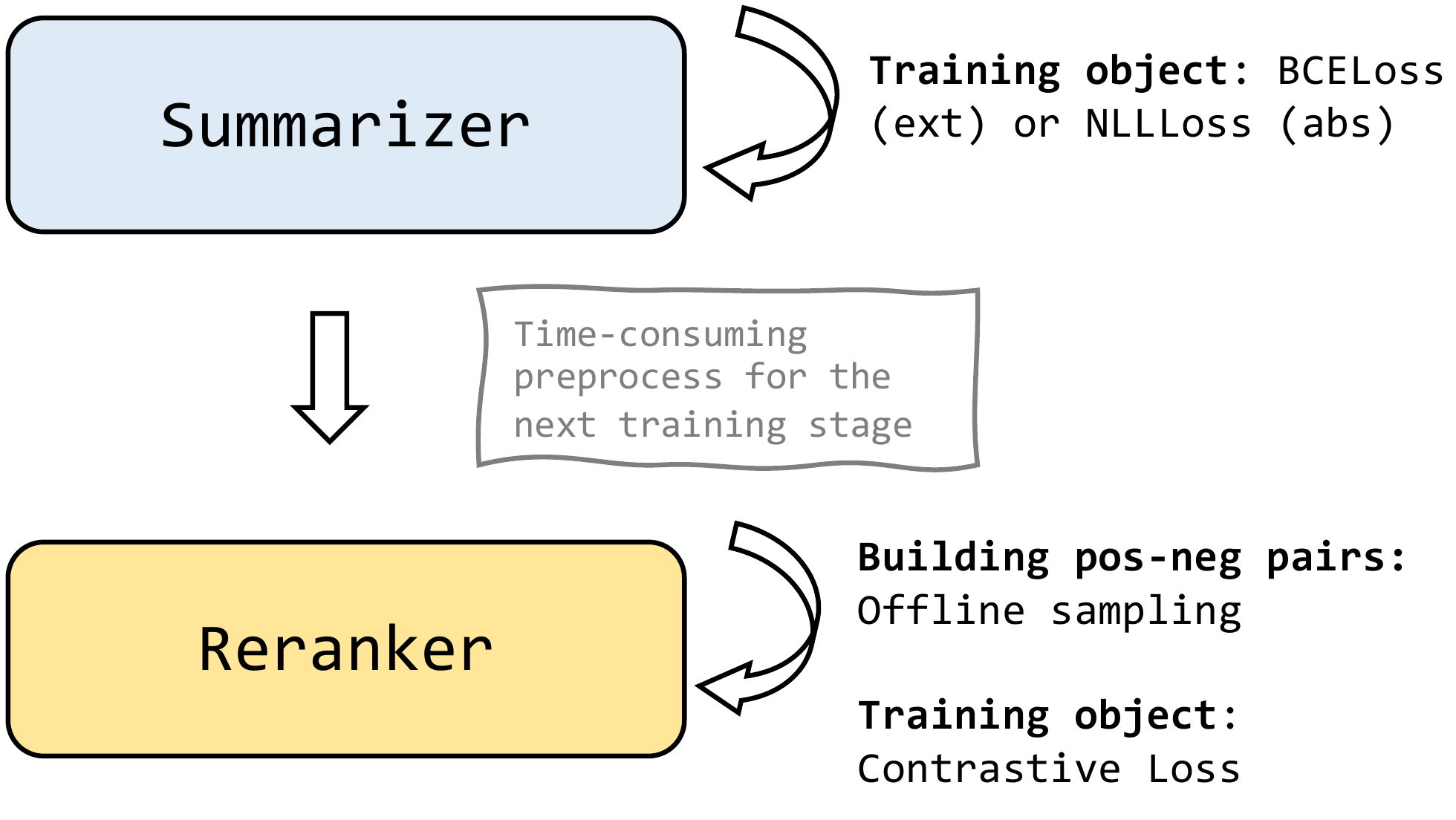}
  }
  \rulesep
  \subfigure[CoLo (this work)]{
    \label{fig:CoLo}
    \includegraphics[width=0.26\textwidth]{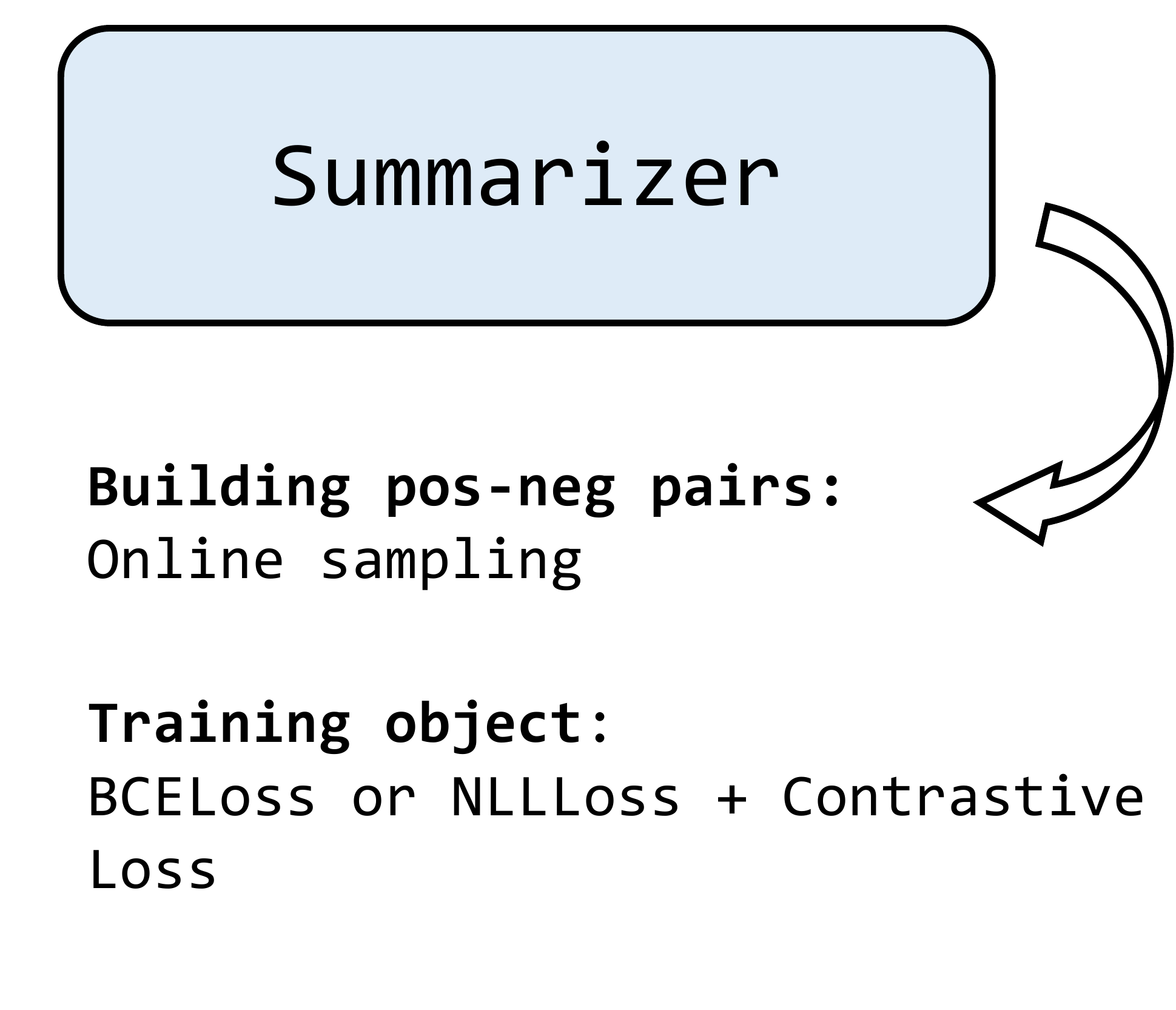}
  }
 \caption{ A comparison between two-stage models and \textsc{CoLo}. The two-stage models including two training stages and a time-consuming preprocess while \textsc{CoLo} is trained in an end-to-end fashion. (GPU and CPU hours cost in each stage are shown in Table~\ref{tab:training effi}). Two-stage models use offline sampling to build positive-negative pairs while  \textsc{CoLo} builds positive-negative pairs with online sampling where we directly get theses pairs from a changing model distribution. }
 \label{fig:compare}
\end{figure*}

\section{Method}
\subsection{A Naive One-Stage Re-ranking Model}
The goal of one-stage re-ranking systems is to enable both training and inference to score candidate summaries holistically without requiring a second stage of computation by a separate model.
Ideally, an one-stage summarization model should both function as  a  summarizer and a re-ranker. A  straightforward solution is multi-task learning.  The naive training pipeline can be formulated as follows: (i) tuning $\mathcal{M}$ with $\mathcal{L}_{sum}$. (ii) Getting positive and negative samples from $p_{\mathcal{M}}(Y|X)$ via offline sampling for each datapoint $X$ in the training set. (iii) Building the ranking loss with these candidates and further tuning $\mathcal{M}$ with $\mathcal{L}_{rank} + \mathcal{L}_{sum}$. However, in practice, such training method is always suboptimal compared to the state-of-the-art two-stage models. We denote the model after multi-task learning as $\mathcal{M'}$. There is a serious \textit{generalization  error} in the naive methods: via multi-task learning, $\mathcal{M}'$ is only able to rank candidates drawn from the original model distribution  $p_{\mathcal{M}}(Y|X)$ but not candidates from the new distribution $p_{\mathcal{M'}}(Y|X)$. This error makes the naive approach unable to directly output a good summary in sequence-level generated by itself.

\subsection{Our approach: \textsc{CoLo}}\label{ext_detail}
The first step of CoLo is also to train the summarization model with $\mathcal{L}_{sum}$ like the naive approach.

In CoLo, we discard using positive-negative samples that from a fixed model distribution, instead, we sample these candidates from a constantly shifting model distribution during multi-task learning.  By doing so, we can mitigate the above mentioned generalization error as much as possible because candidates are dynamically changing with the parameters of the model distribution $p_{\mathcal{M}}(Y|X)$ updated by gradient descent.
To implement this process, at each training step, we sample the newest  candidates along with their feature presentations from the summarization model and calculate the ranking loss. 
We will give a detailed description about how we performing the online sampling process on  mainstream extractive and abstractive summarization models in the following parts. 

\begin{figure}[t]
    \centering
    \includegraphics[width=0.9\linewidth]{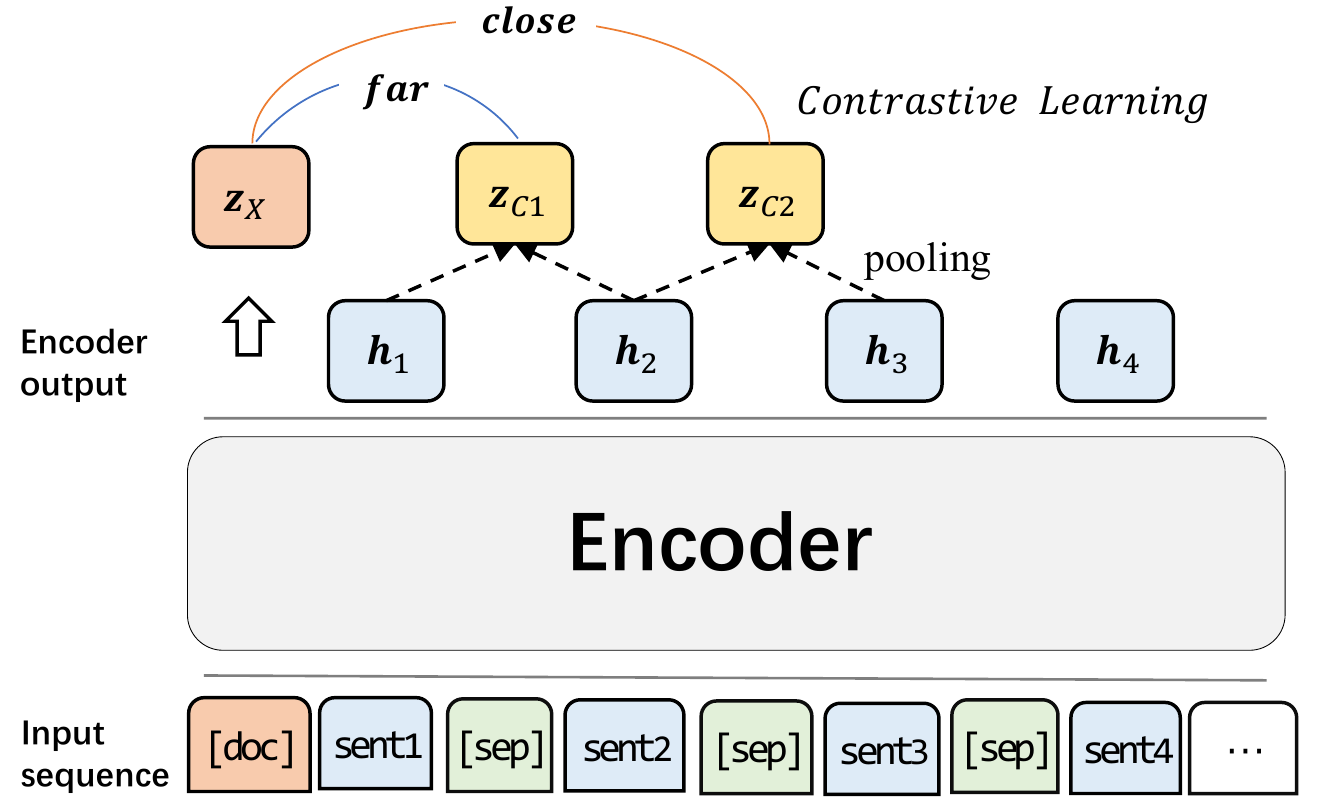}
        \caption{Architecture of our extractive model. Input sequence: The `[doc]' token is used to get vector representation  $\mathbf{z}_{X}$ of the document $X$, `[sep]' is used as separator for sentences. We omit the classifier and the BCELoss. $\mathbf{h}_i$ is the sentence embedding the i-$th$ sentence in $X$.  $\mathbf{z}_{C_i}$ means the feature representation of the $i$-th candidate.}
    \label{fig:ext_model}
\end{figure}

\paragraph{Online Sampling for Extractive Model}
The task of extractive summarization is to assign a label $y_i \in \{0,1\}$  for each sentence ${sent}_{i}$ from the source document  $X$ $ = ({sent}_{1}, {sent}_{2},\ldots,{sent}_{n})$ consisting of $n$ sentences.
Figure~\ref{fig:ext_model} gives an example of our one-stage extractive summarization model. Extractive candidates can be viewed as a subset of sentences from the document.
In this figure, we sample $sent_1, sent_2$ to form the first candidate $C_1 = \{sent_1, sent_2\}$,  and $C_2$ is consisting of $\{sent_2, sent_3\}$. 
After constructing these candidates, the next step is  to represent them in the embedding space. In our one-stage model, we employ a heuristic way to  obtain the feature presentations of candidates: pooling results of the sentence embedding from the extractive model. Concretely, we denote the sentence embedding for the $i$-th sentence as $\mathbf{h}_i$. The hidden representation of a candidate is created by pooling the sentence representations belong to it. For example $\mathbf{z}_{C_1}$ is the average pooling result of $\mathbf{h}_1$ and $\mathbf{h}_2$. Suppose $C_2$ has higher ROUGE score than $C_1$, then $C_2$ is treated as a positive sample and $C_1$ is treated as a negative sample for this pair. Finally, the whole system is trained by the sum of  $\mathcal{L}_{rank}$ and $\mathcal{L}_{sum}$.

Sampling informative candidates is essential in re-ranking systems.
The first step of the sampling method is to determine ${\mathcal{N}}$ which represents the number of candidate sentences. ${\mathcal{N}}$ is set depending on the number of summary sentences of downstream datasets. Take CNN/DailyMail as an example, we set ${\mathcal{N}}$ to $\{2,3\}$ because most gold summaries consist of 2$\sim$3 sentences. At each training step, we iterate over $\mathcal{N}$  by combination and form $m$ different candidates $\mathcal{C} = \{C_1, C_2, \ldots, C_m\}$. $ m$ is equal to $\sum_iC_{n}^{num_i}$  where $num_i$ is the $i$-th element in ${\mathcal{N}}$ and $n$ is number of sentences of the document. For CNN/DailyMail whose ${\mathcal{N}}$ is set to $\{2,3\}$, we can sample $C_n^2 + C_n^3$ different candidates.  

However, in practice, we always face the combination explosion problem when the number of sentences $n$ grows larger.
The two-stage system~\cite{DBLP:conf/acl/ZhongLCWQH20} pre-trained an extractive model to clip the origin number of sentences to an acceptable size. 
Notice that our extractive summarizer is also supervised with the BCELoss, so that we can clip the sampling space to  $n'$ (a hyperparameter) with the output distribution over the sentences  at each training step.  Then the total size of the final candidate set decreases to $m' = \sum_iC_{n'}^{num_i}$. For CNN/DailyMail, $n'$ is set to 5, and we can get $C_5^2 + C_5^3 = 20 $ different extractive candidates.  Details about the setting of $\mathcal{N}$ and $n'$ can be found in Table~\ref{tab:candidate_size} in Appendix. 

Notably, the offline sampling needs to feed each candidate into the pre-trained encoder. In real-life setting, when summarizing some long documents, the number of sentences in the input document and output summary will increase significantly. It will bring a polynomial level increase to the computation and GPU overhead of the two-stage model. But our one-stage system with online sampling is robust to the length variance.

\paragraph{Inference Stage of Extractive Model}
Since we have modeled a summary-level score during training, it is easy to directly generate summaries according to the summary-level semantic score.  Concretely,  given a candidate set $\mathcal{C}$ built by the combination  strategy, we  calculate the cosine similarity between each candidate presentation $\mathbf{z}_{C_i}$ and the document representation $\mathbf{z}_X$:
\begin{equation}
\label{eq:ext_inf}
    \hat{C} =  \max_{C_i \in \mathcal{C}}{cos( \mathbf{z}_{X},\mathbf{z}_{C_i})}.
\end{equation} The final output is the candidate with highest cosine similarity score.

\paragraph{Online Sampling for Abstractive Model}
Our method can also be easily adapted in abstractive summarization.
Selecting a generated summary maximum a posteriori (MAP) usually result in poor performance~\cite{stahlberg2019nmt}, thus most state-of-the-art generation model usually use the beam search algorithm at inference stage. 
The online sampling for the abstractive version is much simpler than the extractive version. We use beam search as sampling algorithm and get the feature representations from the encoder/decoder output.
We denote the encoder output of source document $X$ as $H_{enc}$ and the decoder hidden states of the target summary as $H_{dec}$. We get the document representation from the encoder output of the 0-th token $\mathbf{z}_{X} = H_{enc}^0$. The feature representation of the $i$-th candidate $C_i$ with length = $|C_i|$ is derived from the last step of the decoder output $\mathbf{z}_{C_i} = H_{dec}^{|C_i|-1}$. Hidden states of other steps can not represent the entire sequence because of the sequence mask in transformer decoder. finally we formulate the ranking loss following Eq.~\ref{eq:loss_rank}.

\paragraph{Inference Stage of Abstractive Model}
The inference stage of our abstractive version is similar to the extractive version. We save the feature representation of the document and each beam during beam search.  The final output is determind by the cosine distance between $\mathbf{z}_{X}$ and $\mathbf{z}_{C_i}$.

\section{Experimental Setup}
\renewcommand\arraystretch{1.4}
\begin{table}[t]\footnotesize\setlength{\tabcolsep}{2.3pt}
  \centering
    \begin{tabular}{lccccc}
    \toprule
    \qquad &\textbf{CNN/DM} &  \textbf{Reddit} & \textbf{XSum} & \textbf{SSN} & \textbf{PubMed} \\
    \midrule
    $n'$ & 5 & 5 & 5 & 8 & 8  \\
    $ {\mathcal{N}} $ & 2,3 & 1,2 & 1,2 & 6 & 6,7   \\
    $|\mathcal{C}$| & 20 & 15 & 15  & 28 & 36  \\
    \bottomrule
    \end{tabular}%
  \caption{candidate size $|\mathcal{C}|$ of each datasets (extractive). $|n'|$ is the clipped candidate size, ${\mathcal{N}}$ is a set containing all number of possible sentence.}
  \label{tab:candidate_size}
\end{table}

\subsection{Datasets}
We conduct experiments on five mainstream datasets to evaluate the effectiveness of our approach. \\
\textbf{CNN/DailyMail} \cite{hermann2015teaching} is a classic benchmark  which contains articles from the CNN/Daily Mail newspapers. We use the cased version from datasets\footnote{\url{https://github.com/huggingface/datasets}} \\
\textbf{XSum}~\cite{narayan2018don} is a one-sentence summary dataset from BBC News. Gold summaries are professionally written by the authors of documents.\\
\textbf{Reddit}~\cite{kim2019abstractive} is collected from social media platform and we use the TIFU-long version.\\
\textbf{PubMed}~\cite{cohan2018discourse} is a long document summarization dataset from scientific domain  whose $avg$ summary length is about 4 times longer than CNN/DM.\\
\textbf{SSN}~\cite{an2021enhancing} consists of papers mainly from math, physics and computer science with the abstract section as gold reference.\\

\subsection{Implementation Details}
For the simplity of experimental settings, both extractive model and abstractive mode are based on BART.
We use the encoder of BART (170M) as the backbone and a 3-layer MLP as the classifier to implement the extractor. We add two special token `<cls>' to generate the sentence representation and `<sep>' as sentence separator. `<doc>' token is used to generate the document feature representation. candidate size for each dataset can be found in~\ref{tab:candidate_size}  We use adam optimizer~\cite{kingma2014adam} learning rate schedule follows the setting in transformer~\cite{vaswani2017attention}. We train our model for 15000 steps with BCELoss and 32000 steps with BCELoss and RankingLoss where each step has a batch size of 36. The margin parameter $\gamma$ is set to 0.01. The size of generated candidates $|\mathcal{C}|$ is set to 20 for CNN/DM. We report the results. Other settings follow the default setting in \citet{liu2019text}. Our model is trained on single GeForce RTX 3090 GPU for 8 hours. Both our abstractive model and extractive model are trained on 24G GeForce RTX 3090 GPUs and the inference process is on 12G GeForce GTX TITAN XP GPUs.

For abstractive model, we choose BART initialized with facebook/bart-large-cnn from transformers\footnote{\url{https://github.com/huggingface/transformers}} as the basic summarizer. We further fintune this model by NLLLoss and RankingLoss for 15000 steps where each step with a batch size of 8. Other setting is the same with our extractive version. To encourage diversity, we use the diverse beam search~\cite{vijayakumar2016diverse} to generate the candidates with beam size set to 16 and diversity penalty set to 1.0. Our model is trained on 8 GeForce RTX 3090 GPUs for about 18 hours.

\subsection{Evaluation Metrics}
We examine our approach with 4 metrics that measure the distance between generated summaries against the gold reference. \textbf{ROUGE}~\cite{lin2004rouge} where R-1 and R-2 measure  informativeness based on n-gram overlapping and R-L represents fluency. \textbf{JS-2 Divergence}~\cite{louis2013automatically} measures Jensen-Shannon divergence between the bigram distributions of two input texts. \textbf{BERTScore}~\cite{zhang2019bertscore}  measures soft overlap between BERT embeddings of two texts instead of using lexical matching methods. \textbf{MoverScore}~\cite{zhao2019moverscore} is also based on the neural model but applies a earth mover distance measure to contextualized BERT embeddings.

\section{Results}
We denote the model without contrastive learning as the baseline system. Since the backbone of our extractive model is BART encoder so that we call the baseline model \textsc{BartExt}. The baseline model for abstractive system is BART. Our extractive model is called \textsc{CoLo}$_{Ext}$ and its abstractive version is denoted as \textsc{CoLo}$_{Abs}$.

\subsection{Extractive Results}
We compare our models with baseline models which has similar amount of parameters and decoding speed of our models in this section.
Our extractive results on CNN/DM are shown in Table~\ref{table:ext_cnndm}
We compare our model with previous strong extractive baseline built on pre-trained model~\cite{zhong2019searching, bae2019summary, liu2019text} and strong multi-stage systems~\cite{DBLP:conf/acl/ZhongLCWQH20}. From the third section of Table~\ref{table:ext_cnndm}, we can see that our model \textsc{CoLo}$_{Ext}$ beats the baseline model by 1.49 ROUGE-1 score and achieve the state-of-the-art among all end-to-end systems when input length set to 512 and the results can be further improved while extending the input length to 1024. Even compared with the \textsc{BertSum}-large (340M)~\cite{liu2019text} which is built on large PTM,  We still have an improvement of 0.42  with only the half number of parameters of theirs. Though RL-based methods hold the motivation of optimizing towards the evaluation metric, but it does not gain much improvement on performance in practice.

To verify whether our model is effective on datasets of various lengths, we also evaluate our model on datasets with short summaries (Reddit and XSum) and long document dataset PubMed and results are shown in  Table~\ref{tab:ext_other_datasets}. On reddit and XSum, we achieve the advantage of more than 1.0 point ROUGE-1 than baseline systems and close performance with the upper bound ORACLE. We also gain improvements when tested on the long document summarzation dataset PubMed. Detailed results on long document dataset can be found in Appendix~\ref{sec:long_doc}.
\renewcommand\arraystretch{1.2}
\begin{table}[t]
\center \footnotesize
\setlength{\tabcolsep}{1.05mm}{
\begin{tabular}{lccc}
\toprule
{\textbf{Model}} & \textbf{R-1} & \textbf{R-2} & \textbf{R-L}  \\
\midrule
LEAD & 40.43 & 17.62 & 36.67 \\
ORACLE & 52.59 & 31.23 & 48.87 \\
\midrule
Transformer\cite{vaswani2017attention} & 40.90 &18.02 &37.17 \\
\textsc{Bert-Ext}\cite{bae2019summary} & 42.29 & 19.38 & 38.63 \\
\textsc{Bert-Ext} + RL  & 42.76 & 19.87 & 39.11 \\
{BertSum}~\cite{liu2019text} & 42.57 & 19.96 & 39.04 \\
{BertSum}-large & \underline{43.85} & 20.34 & 39.90 \\
\midrule
\textsc{BartExt} & 42.78 & 20.24 & 39.24 \\
\textsc{BartExt} ($len=1024$) & 43.65 &  \underline{20.88} &  \underline{40.19} \\
\midrule
Naive one-stage  & 43.53  & 20.54 & 39.62 \\
\textsc{CoLo}$_{Ext}$  & 44.10  & 20.97 & 40.19 \\
\textsc{CoLo}$_{Ext}$ + BERTScore & 44.27 & 21.01 & 40.34 \\
\textsc{CoLo}$_{Ext}$ ($len=1024$) & \textbf{44.58} & \textbf{21.25} & \textbf{40.65} \\
\bottomrule
\end{tabular}}
\caption{Extractive results on CNN/DM test set. $len$ means the input length of the document, results without the marker using 512 tokens as input. +RL means the addition of reinforcement learning. +BERTScore means we use BERTScore to determine positive-negative samples. \textsc{CoLo} clearly outperform all previous one-stage summarization systems. The best results are in bold and the second best ones are underlined. }
\label{table:ext_cnndm}
\end{table}

\renewcommand\arraystretch{1.0}
\begin{table*}[t]
\center \footnotesize
\tabcolsep0.13 in
\begin{tabular}{lccccccccc}
\toprule
\multicolumn{1}{c}{\multirow{2}[1]{*}{\textbf{Model}}}  &
\multicolumn{3}{c}{\textbf{Reddit}} &
\multicolumn{3}{c}{\textbf{XSum}} &
\multicolumn{3}{c}{\textbf{PubMed}} \\
 & \textbf{R-1} & \textbf{R-2} & \textbf{R-L} &
\textbf{R-1} & \textbf{R-2} & \textbf{R-L} &
\textbf{R-1} & \textbf{R-2} & \textbf{R-L} \\

\cmidrule(lr){1-1} \cmidrule(lr){2-4} \cmidrule(lr){5-7} \cmidrule(lr){8-10}
LEAD & 12.38 & 2.17 & 10.12 & 14.40 & 1.46 & 10.59 & 37.58 & 12.22. & 33.44 \\
ORACLE & 29.10 & 11.08 & 23.10 & 25.62 & 7.62 & 18.72 & 45.12 & 20.33 & 40.19  \\
\cmidrule(lr){1-1} \cmidrule(lr){2-4} \cmidrule(lr){5-7} \cmidrule(lr){8-10}
\textsc{BertSum} & 23.86 & 5.85	& 19.11 & 22.86 & 4.48	& 17.16 & 41.05 & 14.88 & 36.57 \\

\textsc{BartExt} & 23.97 & 5.68 & 19.24 & 22.96 & 4.70 & 17.29 &41.40 & 16.18 & 37.89 \\
\rowcolor{mygray}
\textsc{CoLo}$_{Ext}$ & \textbf{25.06} & \textbf{5.90} & \textbf{19.52} & \textbf{24.51} & \textbf{5.04} & \textbf{18.21} & \textbf{41.93} & \textbf{16.51} & \textbf{38.28} \\
\bottomrule
\end{tabular}
\caption{Results on test sets of reddit, XSum and PubMed. Our model achieve  significant improvement on the baseline model \textsc{BartExt}. LEAD  means we select the first $k$ sentences from the source document as the output summary and ORACLE is the upper bound of extractive methods. }
\label{tab:ext_other_datasets}
\end{table*}

\subsection{Abstractive results}
Early work also successfully applies reinforcement learning on abstractive summarization~\cite{paulus2017deep,li2019deep}. But we do not find related works that successfully combine reinforcement learning with strong pre-trained models. Therefore, most of our baselines are strong pertrained model finetuned with NLLLoss. Our results is shown in Table~\ref{table:abs_cnndm}, due to the huge cost of using large pre-trained model with length set to 1024, we also report results with 512 input tokens and it is able to significantly outperform other baselines which has longer input length (1024). \textsc{CoLo}$_{Abs}$ has an improvement of \textbf{2.17} R-1 socre on the very strong BART-large baseline without adding additional parameters or modules. Additionally, our method is able to outperform all one-stage baseline systems by a large margin. We also conduct experiments on long document summarzation datasets (see in Table~\ref{tab:abs_long} in Appendix).

\renewcommand\arraystretch{1.3}
\begin{table}[t]
\center \footnotesize
\tabcolsep0.07in
\setlength{\tabcolsep}{0.75mm}{
\begin{tabular}{lccc}
\toprule
{\textbf{Model}} & \textbf{R-1} & \textbf{R-2} & \textbf{R-L}  \\
\midrule
{BertSumAbs}\cite{liu2019text} & 41.72 &19.39 &38.76 \\
Pegasus\cite{zhang2020pegasus} & 44.17 & 21.47& 41.11 \\
BART\cite{lewis2020bart} & 44.16 & 21.28& 40.90 \\
BART+R3F\cite{aghajanyan2020better} & {44.38} & {21.53} & {41.17} \\
BART ($len=512$) & 43.82 & 20.96 & 40.63 \\
\midrule
ConSum~\cite{sun2021alleviating} & 44.53 & 21.54 & 41.57 \\
SeqCo~\cite{xu2021sequence} &\underline{45.02} & \underline{21.80} & \underline{41.75}\\
\midrule
Naive one-stage (ROUGE, $len=512$) & 43.90 & 20.88 & 40.69 \\
\textsc{CoLo}$_{Abs}$ (ROUGE, $len=512$)& 45.45 & 21.53 & 42.35 \\
\textsc{CoLo}$_{Abs}$(ROUGE) & \textbf{46.33} & \textbf{22.15} & \textbf{43.08} \\
\bottomrule
\end{tabular}}
\caption{Abstractive results on CNN/DM test set. $len$ means the maximum input length of the encoder, results without the marker using 1024 tokens as the input. ConSum~\cite{sun2021alleviating} and SeqCo~\cite{xu2021sequence} in the second block are also previous contrastive learning based methods without re-ranking. }
\label{table:abs_cnndm}
\end{table}

\subsection{Comparison with Multi-stage Systems}\label{sec:effi}
Apart from the one-stage systems,  we also compare our model with these powerful multi-stage systems: CTRLSum, multi-stage re-ranking models.  CTRLSum needs  other systems to previously produce a control signal.

\renewcommand\arraystretch{1.2}
\begin{table}[t]
\center \footnotesize
\setlength{\tabcolsep}{0.5mm}{
\begin{tabular}{lccc}
\toprule
{\textbf{Model}} & \textbf{R-1} & \textbf{R-2} & \textbf{R-L}  \\
\midrule
\multicolumn{4}{c}{\textit{extractive systems}} \\
\midrule
\textsc{CoLo}$_{Ext}$ & 44.27 & 21.01 & 40.34 \\
BERT+BERT$^\mathcal{R}$~\cite{DBLP:conf/acl/ZhongLCWQH20} & 44.22 & 20.62 & 40.38 \\
BERT+RoBERTa$^\mathcal{R}$ ~\cite{DBLP:conf/acl/ZhongLCWQH20}  & 44.41 & 20.86  & 40.55\\
\textsc{CoLo}$_{Ext}$ +RoBERTa$^\mathcal{R}$ & \textbf{44.70} & \textbf{21.03}  & \textbf{40.74}\\
\midrule
\multicolumn{4}{c}{\textit{abstractive systems}} \\
\midrule
\textsc{CoLo}$_{Abs}$ & 46.33 & 22.15 &  43.08 \\
{CTRLSum}\cite{he2020ctrlsum} & 45.65 & \textbf{22.35} & 42.50\\
BART+RoBERTa$^\mathcal{R}$~\cite{liu2021simcls} & \textbf{46.67} & 22.15 & \textbf{43.54} \\
\bottomrule
\end{tabular}}
\caption{ Comparision with the multi-stage systems. RoBERTa$^\mathcal{R}$ means a RoBERTa re-ranker is is added to the summarization model. }
\label{table:multi-stage}
\end{table}

\paragraph{performance}   The addition of another pre-trained model implicitly introduces more parameters and knowledge, thus  it is usually unfair to directly compare one-stage systems with the two-stage systems. But we show that \textsc{CoLo} is able to achieve comparable performance with the multi-stage systems.  As is shown in the first part of Table~\ref{table:multi-stage},
compared with the multi-stage models that ensembles another pre-trained encoder as a re-ranker, \textsc{CoLo}$_{Ext}$ still performs better than their BERT+BERT$^\mathcal{R}$ version without the need to re-feed the generated candidates to another model meanwhile we obtain a \textasciitilde$\times 5$ speed up over the multi-stage systems. We also try concatenating a re-ranker RoBERTa for our model, results shows that \textsc{CoLo}$_{Ext}$ can be further improved by combing another pre-trained re-ranker reaching new extractive SOTA on the test set of CNN/DM. For abstractive models,  our end-to-end model still legs behind multi-stage systems but we do not need training another model and keep similar inference speed with baseline models.

\begin{figure}[ht!]
  \centering

  \subfigure[CNN/DM (batch=1)]{
    \label{fig:cnn1}
    \includegraphics[width=0.22\textwidth]{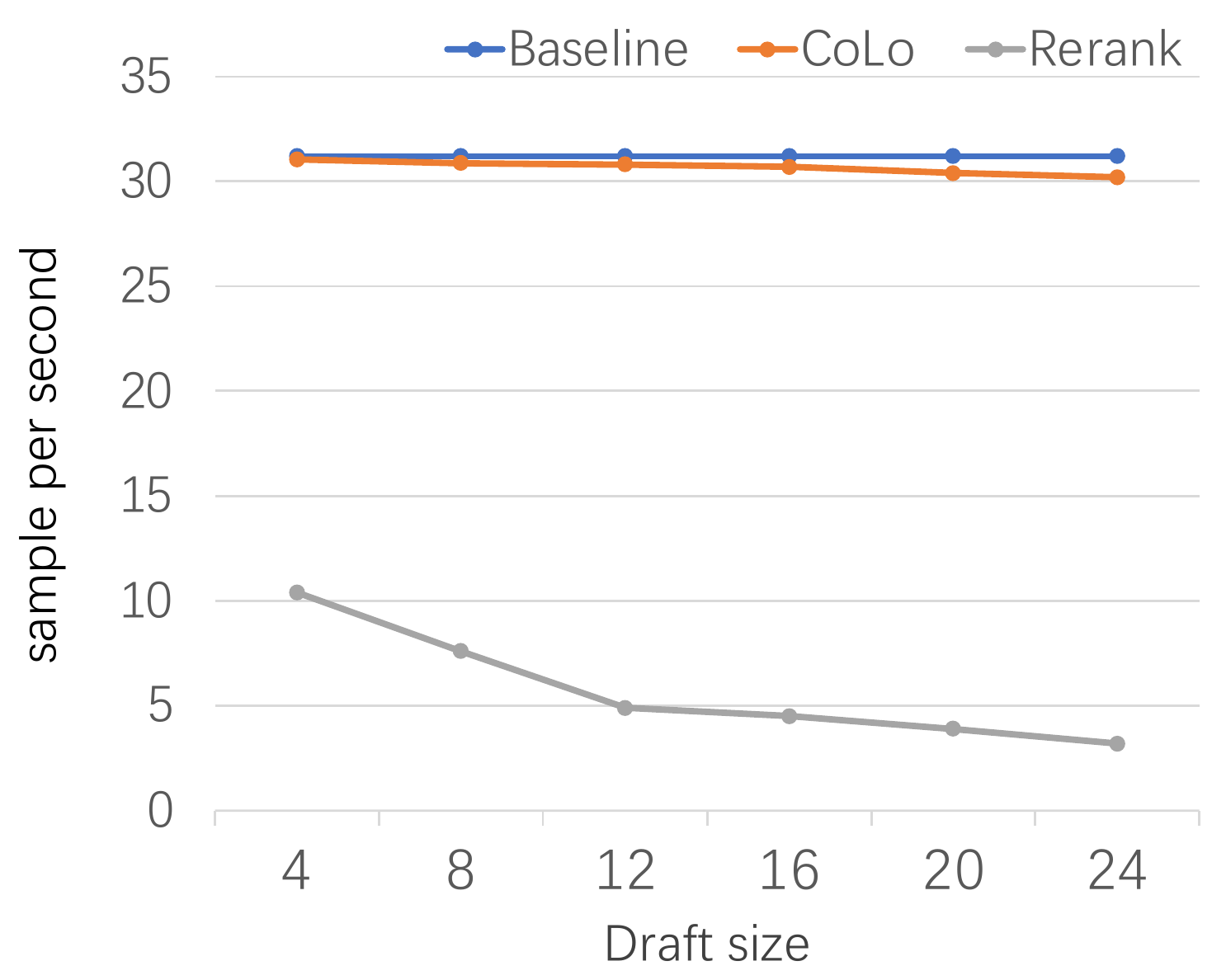}
  }
  \subfigure[CNN/DM (batch =\textit{MAX})]{
    \label{fig:cnnm}
    \includegraphics[width=0.22\textwidth]{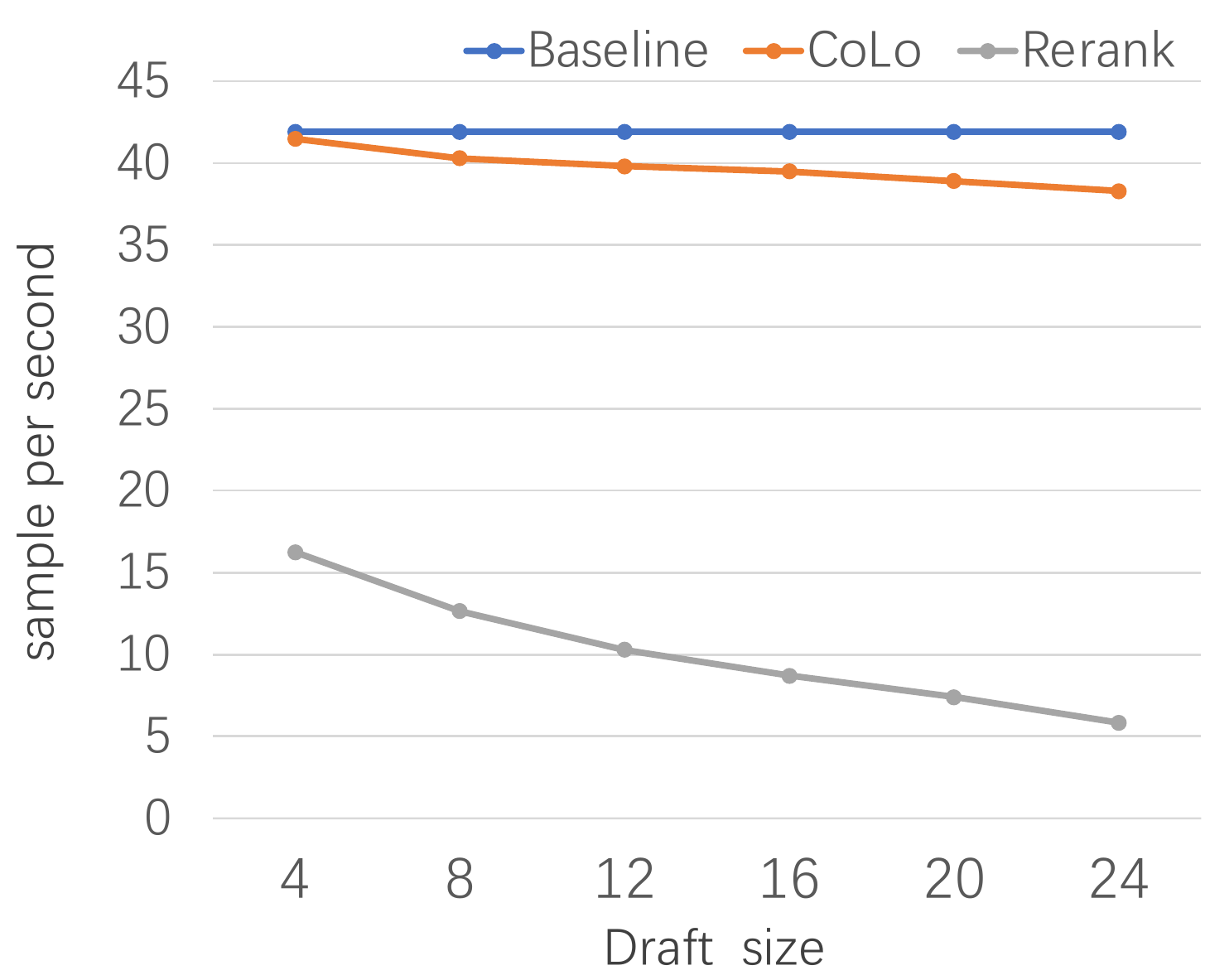}
  }

 \caption{Inference speed on CNN/DM (extractive).  we use the candidate size $|\mathcal{C}|$ as the X-axis. The Y-axis represents the number of samples processed per second. batch=\textit{{MAX}} means we use the maximum batch size allowed by GPU memory. }
 \label{fig:decoding_time}
\end{figure}

\begin{figure}[t]
    \centering
    \includegraphics[width=0.7\linewidth]{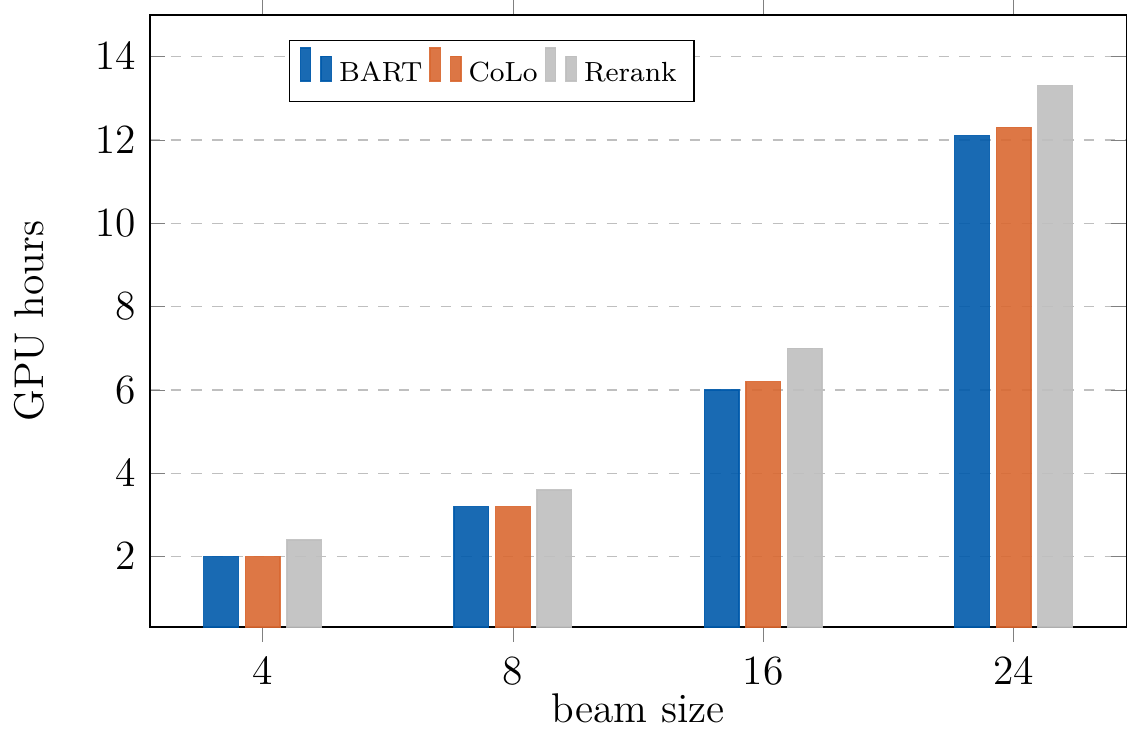}
        \caption{ Test inference time  with beam size for abstractive model. We use the maximum batch size allowed by GPU memory.  }
    \label{fig:abs_time}
\end{figure}

\paragraph{Inference Efficiency} Despite the fact that multi-stage models outperform all end-to-end systems, they frequently suffer from inefficiency. In this part we mainly focus on analysing the efficiency of 3 kinds of systems: 1) baseline, which is trained only with BCELoss or NLLLoss, 2) \textsc{CoLo}, our end-to-end constrastive learning framework, 3)  Rerank, which means the multi-stage re-ranking systems. it has more 110M parameters than baseline model and  \textsc{CoLo}. The efficiency experiments for training and inference are respectively conducted on 24G RTX 3090 GPUs and 12G TITAN XP GPUs. For extractve summarization, figures~\ref{fig:cnn1},\ref{fig:cnnm} give a detailed comparison of the inference speed between the three models. Y-axis  represents the  number  of  samples  processed per second. To give a fair comparison, we test the inference efficiency in two settings: i) all models are tested with batch size fixed to 1. ii) all models are tested with the maximum batch size allowed by the GPU.  While the candidate size varies from 4$\sim$32, both our model have a $\mathbf{3\times \sim 8\times}$ speed-up ratio over the multi-stage re-ranking model. When the candidate size is set to 20, the baseline model is able to process \textasciitilde31.2/41.9 (batch = 1/\textit{MAX}) samples per second, the decoding speed of \textsc{CoLo}$_{Ext}$ is \textasciitilde30.4/38.9 samples/s (batch=1/\textit{MAX}) and the decoding speed of the multi-stage re-ranking model is only \textasciitilde4.9/7.0 samples/s(batch=1/\textit{MAX}). Our model almost does no harm on inference speed while the candidate size $|\mathcal{C}|$ is less than 16. However, when the candidate size grows larger there is more time spent on generating the representations of the candidates. Figure~\ref{fig:abs_time} show the comparison of inference time of  the  abstractive models. While the bottleneck of abstractive models is the auto-regressive generation process. Our abstractive model generally save \textasciitilde0.5 GPU hours compared to the re-ranking model.

\begin{table}[t]\footnotesize\setlength{
\tabcolsep}{1.5pt}
  \centering
    \begin{tabular}{lcccc}
    \toprule
    \textbf{Systems} &\textbf{Stage1} & \textbf{Preprocess} & \textbf{Stage2} & \textbf{Total hours}\\
    \midrule
    Ext+RoBERTa$^\mathcal{R}$  & 4 & 5 (+20) & 128 & 137 (+20)  \\
    \textsc{CoLo}$_{Ext}$  & 7 & -- & -- & 7 ($\downarrow\mathbf{130}$)  \\
    \midrule
    Abs+RoBERTa$^\mathcal{R}$ & 80 &  132 (+18) & 128 & 340 (+18) \\
    \textsc{CoLo}$_{Abs}$ & 224 & -- & -- & 224 ($\downarrow\mathbf{116}$)  \\
    \bottomrule
    \end{tabular}%
  \caption{ GPU hours spent on training for each process on the training set of CNN/DM(reported results are rounded down after the decimal point.  Ext+RoBERTa$^\mathcal{R}$/Abs+RoBERTa$^\mathcal{R}$ denotes the multi-stage re-ranking systems with an extracitve/abstrastive summarizer. (+18)/(+20) means 18/20 CPU hours are spent on calculate ROUGE score for each candidate with 32 threads. }
  \label{tab:training effi}
\end{table}

\paragraph{Training Efficiency} Table~\ref{tab:training effi} gives an overview of the training time of our system and the  multi-stage models on the training set of CNN/DM. The general pipeline for the multi-stage models is: i) training a generator {(Stage1)}, ii) {Preprocess}, ii) training a re-ranker  (Stage2).  The preprocess includes generating the training/dev/test set for training re-ranker and sorting candidates by ROUGE. For extractive system we save \textbf{130} GPU hours compared to the multi-stage systems whose bottleneck is training the re-ranking model. For abstractive model, apart from the 128 GPU hours spent on training the ranker, using beam search to generate the training set for re-ranker model is also very time consuming,  generally we obtain \textbf{116} GPU hours and 18 CPU hours saved.

\subsection{Ablation for Different Discriminators} 
In addition to ROUGE, we also select other metrics as the discriminator (shown in Table~\ref{table:cmp_metrics}). ROUGE and JS-2 is based on lexical matching while BERTScore and MoverScore are based on the contextualized embedding from BERT. Our model generally obtains the best results on the metric used in training. Because these metrics are not actually separated, using one of these metrics as the discriminator can also gain significant improvements on other metrics. Overall, the neural evaluation metric BERTScore and MoverScore bring more improvements compared with metrics that based on the lexical matching. But incorporating neural model based metrics in training will obviously increase the training time.

\renewcommand\arraystretch{1.2}
\begin{table}[t]
\center \footnotesize
\tabcolsep0.07in

\setlength{\tabcolsep}{1.2mm}{
\begin{tabular}{lcccccc}
\toprule
{\textbf{Metric Used}} & \textbf{R-1} & \textbf{R-2} & \textbf{R-L}& \textbf{JS-2} & \textbf{BS} & \textbf{MS}  \\
\midrule
Baseline & 42.78 & 20.24 & 39.23 & 54.24 & 43.52 & 58.27 \\
ROUGE-1,2 &  \best 44.10  & \best 20.97 & 40.19 & 54.07 & 44.26 & 58.63 \\
ROUGE-L & 44.09 & 20.93 & \best \bf40.34 & 54.06 & 44.32 & 58.60 \\
JS-2 & 43.85 & \textbf{21.13} & 39.98 & \best \textbf{53.92} & 44.19 & 58.60 \\
BERTScore & \textbf{44.27} & 21.01 & \textbf{40.34} & 54.08 & \best\textbf{44.85} & 58.71 \\
MoverScore & 44.21 & 20.81 & 40.25 & 54.33 & 44.47 & \best\textbf{58.78} \\

\bottomrule
\end{tabular}}
\caption{Extractive results of using different evaluation metrics as the discriminator on CNN/DM test set. }
\label{table:cmp_metrics}
\end{table}
\subsection{Visualization Experiment} 
\begin{figure}[ht!]
  \centering
    \subfigure{
    \includegraphics[width=0.225\textwidth]{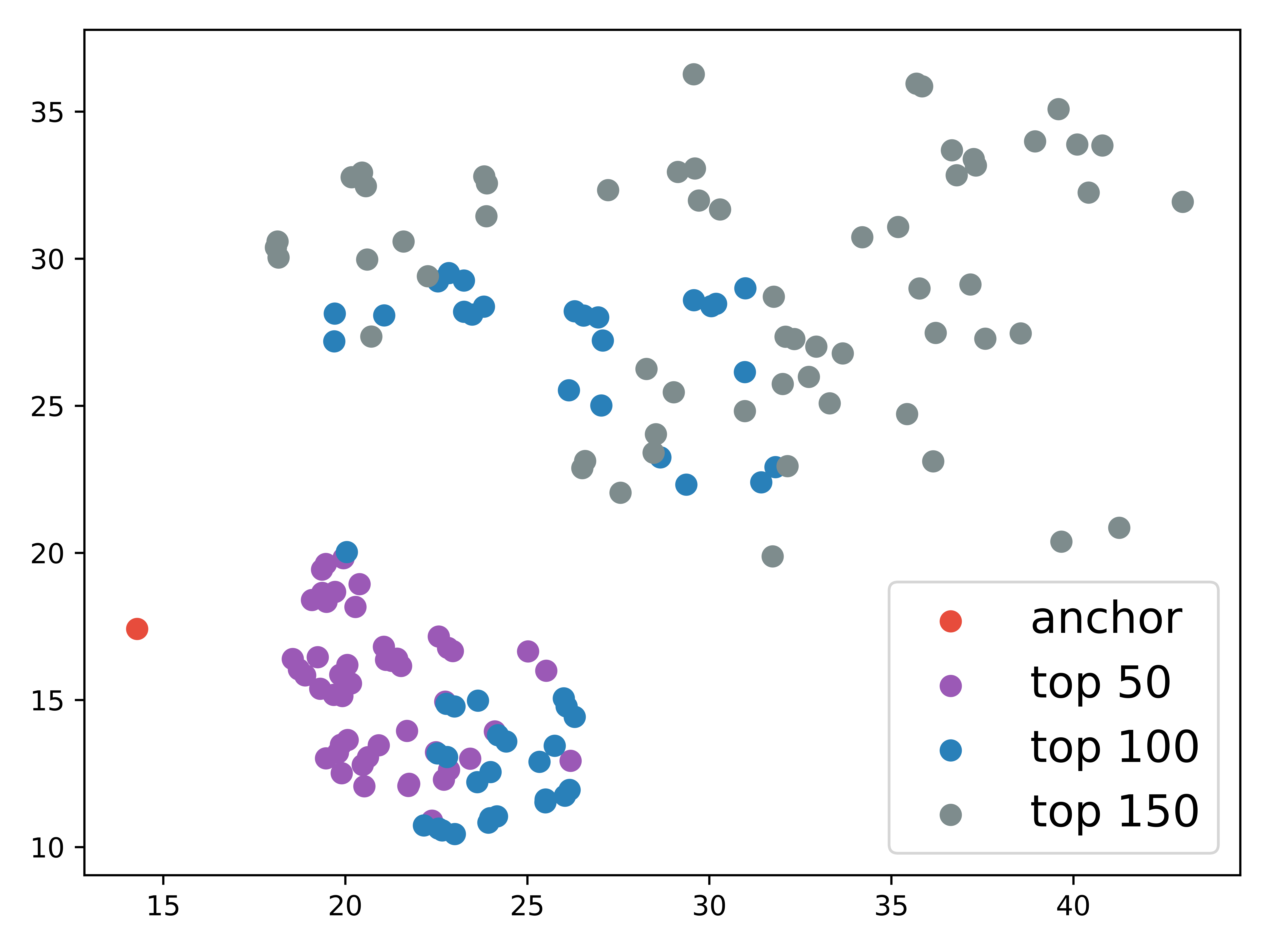}
  }
  \subfigure{
    \includegraphics[width=0.225\textwidth]{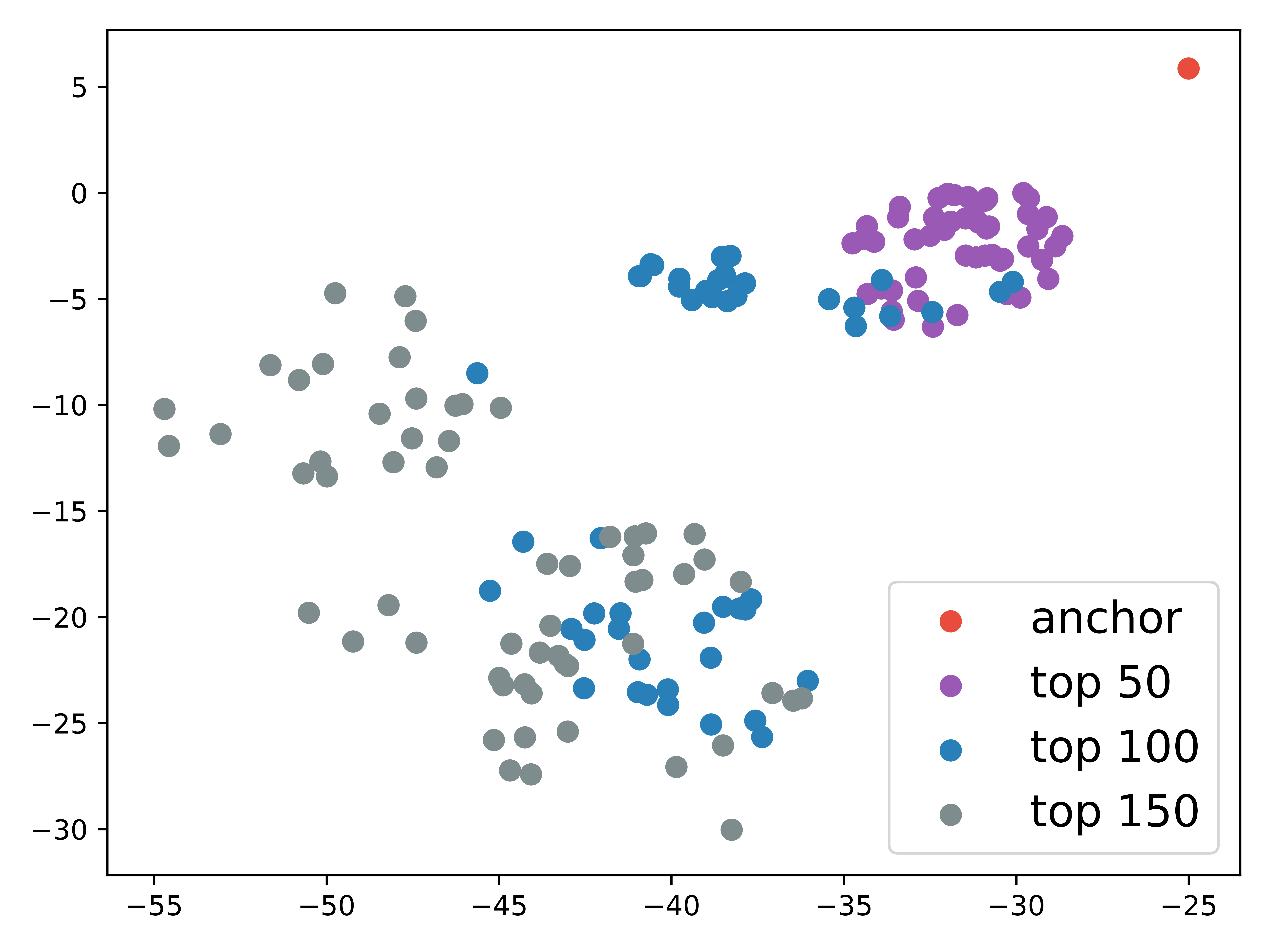}
  }
 \caption{T-SNE Visualization of two examples from CNN/DM test set. We divide the candidates into 3 groups based on ROUGE score: candidates ranking 1\textasciitilde50, candidates ranking 51\textasciitilde100, candidate ranking 101\textasciitilde150. The red point denotes the anchor and the purple/cyan/gray points respectively denote the top 50/100/150 candidates.}
 \label{fig:viz}
\end{figure}

We conduct a visualization experiment on our extractive model to get a close look on the distribution of candidates in semantic space. We randomly sample 100 documents with more than 10 sentences  from the test set of CNN/DM. We first select the top 10 sentences based on the predicted score from the classifier. We set the possible number of sentences to $\{2, 3\}$ resulting a candidate size of $C_{10}^2 + C_{10}^3 = 165$ for each sample.  We visualize the learned embedding of these candidates and the anchor in a two-dimensional space by applying the t-SNE algorithm. As shown in Figure~\ref{fig:viz}, there is an obvious cluster of the top 50 candidates (colored in purple) and the candidates with higher score are closer to the anchor while the distribution of uninformative candidates (gray,cyan points) is relatively random.

\subsection{Human Evaluation}\label{human_eval}
We also conduct a human evaluation on our models to get more accurate results . We randomly select 30 articles from the test set of CNN/DM, and each articles have 5 candidate summaries 4 from automatic systems and 1 is the gold reference. We recruit 2 PhD students majoring in computer science and ask them to rank the candidate summries based on the fluency, informativeness.  If two of these systems generate the same summary for the source document, this sample will be filtered out. As we can see from Table~\ref{tab:human_evaluation}, the \textsc{CoLo}$_{Ext}$ with the discriminator as BERTScore achieve the best result among all automatic systems. However, using BERTScore will bring much training time. We also suggest taking  JS-2 divergence as the discriminator which also does a good job in human evaluation.

\renewcommand\arraystretch{1.2}
\begin{table}[!ht]
\center \footnotesize
\tabcolsep0.05in
\setlength{\tabcolsep}{0.9mm}{
\begin{tabular}{lcccccc}
        \toprule
        \textbf{Metric Used}  & \textbf{1st} & \textbf{2nd} & \textbf{3rd} & \textbf{4th} & \textbf{5th} & \textbf{Avg R.} \\
        \midrule
        Baseline & 0\% & 8.3\%& 8.3\% & 23.3\% & 60\% & 4.33\\
        JS2 & 6.7\% & 25\% & 33.3\%& 21.7\% & 13.3\% & 3.10\\
        R1+R2 & 5\% & 20\% & 28.3\% & 30.3\% & 16.7\% & 3.35\\
        BERTScore & 10\% & 35\% & 20\%& 25\% & 10\% & 2.90\\
        Gold label & 78.3\% & 11.7\% & 10\% & 0\% & 0\% & 1.32\\
        \bottomrule
    \end{tabular}}
\caption{ Results of human evaluation results. Baseline means the \textsc{BartExt} model, Gold-label means the means the human written summary. Avg R. denotes the average ranking of the system.    }
\label{tab:human_evaluation}
\end{table}

\section{Limitations and Future Work}
Compared with the most well-known contrastive learning framework simCLR~\cite{chen2020simple} which propose to construct positive and negative pairs from training samples in the same batch, Drawing negative-positive pairs from the summarization model requires more training time. Ideally, providing  more positive and negative samples will benefit the performance of \textsc{CoLo} . However, decoding with very large beam size in training mode will cost more GPU memory and training time. Future work can search for an efficient way to construct these positive-negative pairs to perform re-ranking during training.

\section{Conclusion}
We introduce \textsc{CoLo}, a contrastive learning based summarization framework for one-stage summarization where positive-negative pairs are generated directly from the summaizer with online sampling. \textsc{CoLo} can be both easily applied on extractive and abstractive methods.  Results show that we greatly exceed  previous stage-of-the art one-stage systems with no additional parameters and obivious decline of the inference efficiency. 

\section*{Acknowledgement}
We would like to thank Yixin Liu and the anonymous reviewers
for their valuable advice.
This work was supported by the National Key Research and Development Program of China (No.2020AAA0106702) and National Natural Science Foundation of China (No.62022027).

\bibliography{anthology,custom}
\bibliographystyle{acl_natbib}

\newpage
\qquad
\appendix

\section*{Appendix}

\section{Results on Long Document Summarization}\label{sec:long_doc}
We experiment our method on two long document datasets PubMed and SSN. A crucial problem for the two-stage model is that they face great difficulty when runing on long document summarization datset. For PubMed, which has an average of 7.6 sentences and 260 tokens after converted to ids. If we have a candidate size $|\mathcal{C}|$= $C_7^6$ + $C_8^7$ and limit the maximum candidate length to 300 that will lead to the total input tokens of the re-ranker up to $(300*15)*batch$.  This will cause an out-of-memory problem even the batch size is set to 1 during training on a 12G GPU.

The two-stage model \textsc{Rerank} in Table~\ref{table:mem_analysis} is implemented with the \textsc{BartExt} as generator and RoBERTa as re-ranker and we report the maximum GPU memory footprint of each system. Compared with the baseline model our approach do not need more  GPU memory during both training and reference while the pipeline model  needs 9.7G memory even with a very small candidate size 8. Since our learning approach does not suffer from the out-of-memory problem, we are able to experiment with larger candidate size and obtain more performance improvements.

\renewcommand\arraystretch{1.2}
\begin{table}[!ht]
\center \footnotesize
\tabcolsep0.18in
\setlength{\tabcolsep}{1.4mm}{
\begin{tabular}{lccc}
\toprule
{\textbf{Model}} & $|\mathcal{C}|$ & \textbf{Mem(T)} & \textbf{Mem(I)}  \\
\midrule
\textsc{BartExt}  & -- & 5.5G & 1.5G \\
\midrule
\multicolumn{4}{c}{\textit{multi-stage systems}} \\
\midrule
\textsc{Rerank} & $C_7^6$  & 8.8G & 1.9G \\
\textsc{Rerank} &  $C_8^7$ & 9.7G & 2.1G \\
\textsc{Rerank} & $C_7^6$ + $C_8^7$ & OOM & 2.4G \\
\midrule
\multicolumn{4}{c}{\textit{ours}} \\
\midrule
\textsc{CoLo}$_{Ext}$ &  $C_7^6$  & 5.6G & 1.5G \\
\textsc{CoLo}$_{Ext}$ & $C_8^7$  & 5.6G & 1.5G\\
\textsc{CoLo}$_{Ext}$ &  $C_7^6$ + $C_8^7$  & 5.6G & 1.5G \\
\bottomrule
\end{tabular}}
\caption{GPU memory test on the test set of PubMed. Mem(T)/Mem(I) denotes the maximum GPU memory used during training/inference with batch size set to 1. Our model hardly need more GPU memory. All these experiments are run on single 12G Geforce TITAN XP GPU. }
\label{table:mem_analysis}
\end{table}

As we can see in Table~\ref{tab:ext_long} and \ref{tab:abs_long}, both the extractive and abstractive model outperform the baseline model on long document. Introducing more samples in comparative learning is helpful for performance, previous two-stage framework has the limitation to expand the candidate size due to is huge memory consumption.

\begin{table}[!ht]
\center \footnotesize
\tabcolsep0.08 in
\begin{tabular}{lccc}
\toprule
\multicolumn{1}{c}{\textbf{Model}} & \textbf{R-1} & \textbf{R-2} & \textbf{R-L} \\
\midrule
\multicolumn{4}{c}{\textbf{SSN}} \\
\midrule
ORACLE & 51.04 & 23.34 & 45.88 \\
\textsc{BertSum} & 42.41 & 13.10 & 37.97 \\
\textsc{BartExt}& 43.53 & 13.45 & 38.00 \\
\textsc{CoLo}$_{Ext}$ & \textbf{42.78} & \textbf{14.21} &  \textbf{42.18} \\
\midrule
\multicolumn{4}{c}{\textbf{PubMed}} \\
\midrule
ORACLE & 45.12& 20.33& 40.19\\
\textsc{BartExt} & 41.40 & 16.18 & 37.89 \\
\textsc{Rerank} (two-stage) &  & &  \\
\qquad  $|\mathcal{C}|$= $C_7^6$ &41.80&16.28&38.20 \\
\qquad  $|\mathcal{C}|$=$C_8^7$ & 41.78 & 16.33 & 38.27\\
\textsc{CoLo}$_{Ext}$  &  & &  \\
\qquad  $|\mathcal{C}|$= $C_7^6$ & 41.74 & 16.28 & 38.20 \\
\qquad  $|\mathcal{C}|$=$C_8^7$ & 41.78 & 16.33 & 38.27\\
\qquad  $|\mathcal{C}|$= $C_7^6$ + $C_8^7$ &\textbf{41.93} & \textbf{16.51}& \textbf{38.28} \\
\bottomrule
\end{tabular}
\caption{Extractive results on test sets of long document dataset PubMed and SSN. Introducing more samples in comparative learning is helpful for performance. But two-stage model has difficulty training with large candidate size. } \label{tab:ext_long}
\end{table}

\begin{table}[!ht]
\center \footnotesize
\tabcolsep0.1 in
\begin{tabular}{lccc}
\toprule
\multicolumn{1}{c}{\textbf{Model}} & \textbf{R-1} & \textbf{R-2} & \textbf{R-L} \\
\midrule
\multicolumn{4}{c}{\textbf{SSN}} \\
\midrule
\textsc{BertSumAbs} & 45.23& 16.56 & 41.25 \\
\textsc{Bart}-base & 45.89 & 16.75 & 41.51 \\
\textsc{CoLo}$_{Abs}$(BERTScore) & 46.23 & 17.09 &  42.10 \\
\textsc{CoLo}$_{Abs}$(ROUGE) & \textbf{46.57} & \textbf{17.21} & \textbf{42.18} \\
\midrule
\multicolumn{4}{c}{\textbf{PubMed}} \\
\midrule
\textsc{BertSumAbs} & 42.90 & 17.35 & 38.88 \\
\textsc{Bart}-base & 43.30 & 17.01 & 39.34 \\
\textsc{CoLo}$_{Abs}$(BERTScore)  & 43.63 & 17.32 & 40.01 \\
\textsc{CoLo}$_{Abs}$(ROUGE) & \textbf{43.98} & \textbf{17.39} & \textbf{40.35} \\
\bottomrule
\end{tabular}
\caption{Abstractive results on test sets of long document dataset PubMed and SSN. Due to the generation process for long document is longer } \label{tab:abs_long}
\end{table}

\end{document}